\documentclass[12pt]{article}
\usepackage[utf8]{inputenc}
\usepackage{setspace}
\usepackage{listings}
\usepackage{graphicx}
\usepackage{fullpage}

\title{Implementing Reinforcement Learning Algorithms in Retail Supply Chains with OpenAI Gym Toolkit}
\author{Shaun D'Souza}
\date{}

\begin{document}

\maketitle

\begin{abstract}
From cutting costs to improving customer experience, forecasting is the crux of retail supply chain management (SCM) and the key to better supply chain performance. Several retailers are using AI/ML models to gather datasets and provide forecast guidance in applications such as Cognitive Demand Forecasting, Product End-of-Life, Forecasting, and Demand Integrated Product Flow. Early work in these areas looked at classical algorithms to improve on a gamut of challenges such as network flow and graphs. But the recent disruptions have made it critical for supply chains to have the resiliency to handle unexpected events.  The biggest challenge lies in matching supply with demand. 

Reinforcement Learning (RL) with its ability to train systems to respond to unforeseen environments, is being increasingly adopted in SCM to improve forecast accuracy, solve supply chain optimization challenges, and train systems to respond to unforeseen circumstances. Companies like UPS and Amazon have developed RL algorithms to define winning AI strategies and keep up with rising consumer delivery expectations. While there are many ways to build RL algorithms for supply chain use cases, the OpenAI Gym toolkit is becoming the preferred choice because of the robust framework for event-driven simulations. 

This white paper explores the application of RL in supply chain forecasting and describes how to build suitable RL models and algorithms by using the OpenAI Gym toolkit. 
\end{abstract}


\section{Why RL is Well Suited for Building AI Models for Supply Chains}
 
The biggest challenge in building AI models for supply chains lies in matching supply with demand given the uncertain environment. Earlier attempts at building AI models for forecasting involved using time series modeling by using a combination of statistical and ML models to forecast sales based on the sales history and trends. But there were inherent limitations wherein such models could forecast numeric data but would be unable to determine policy dynamics. 

OpenAI Gym RL algorithms have several advantages that make them suitable for SCM such as the following: \cite{openai, deepq}

\begin{itemize}
\item \textbf{Ability to optimize strategies and handle unexpected scenarios:} With their optimization procedures, RL algorithms can be used to find the best way to build predictive models that learn over time and earn maximum rewards, i.e. optimize strategies for best outcomes. In reinforcement learning, an agent takes an action in the given environment either in continuous or discrete manner to maximize its reward. Rewards are provided time stepwise and RL applications improve their performance by receiving rewards and punishments from the environment, and thereby determine the best action/strategy for handling situations.
\item \textbf{Ability to cater to a diverse set of use cases:} AI models for SCM need to implement both discrete and continuous models because these cater to a diverse set of use cases in an ever-growing customer environment. By using a RL agent in conjunction with an event simulator available in the OpenAI Gym API, both discrete and continuous models can be built. With a robust framework for event-driven simulations delivered through an agent, actions, and a parameterized environment, OpenAI Gym with a Python-based agent work well for SCM AI models.
\item \textbf{Ability to work with both structured and unstructured data:}  Python\footnote{Note: The proliferation of off-the-shelf open source library APIs such as Google TensorFlow and Keras for high performance numerical computation facilitate the development of AI/ML based solutions using an open source development model. And, programming languages serve as an interface to the algorithms and lend a level of extensibility to users (AI experts, Researchers). Amongst them, Python with a large number of open-source libraries is the most popular.} can run sophisticated kernels owing to the availability of the NumPy and SciPy libraries. For example, Pandas, a Python library, provides support for importing and writing data sets in the widely used .xls and .csv formats. It also supports the JSON format making it possible to read a structured dataset with features and iterate through the rows in the data \cite{sklearn}. However, unstructured data has to be parsed and then read using a visitor function. 
\end{itemize}

\section{It is All About Real Data and Business Rules}

AI/ML models learn outcomes from real data/parameters and business rules \cite{rlrailway}. Let's understand how a parameterized RL model can be applied to supply chains with an example of promotional forecasting, i.e., predicting store sales for a given calendar day based on historical sales data. Typically such promotions are governed by retailer goals and constraints/business rules.
 
For example, some of the available data points/parameters in a typical forecasting model are summarized in Listings ~\ref{promotionplan} -~\ref{rxtransactions}. The data points provided are typically evaluated in conjunction with business rules such as seasonality, promotional events, day of the week, and state or school holidays, and they form the basis of the parameterized environment. 

\begin{lstlisting}[caption=Promotion Plan, label=promotionplan]
Promo Code	Promo Type	Event ID	Promo Start Date	Promo End Date	Promo Target/Projection Amount Store ID	Ad ID/Block ID	Product ID	Offer Qty Offer Price	Planogram Change Indicator	Special Package Indicator	Ad Location Indicator	Coupon Indicator
\end{lstlisting}

\begin{lstlisting}[caption=Online Transactions, label=onlinetransactions]
Product ID	Date	EoD Sales Qty	EoD Return Qty	Zip Code	City	State Geo Area Code
\end{lstlisting}

\begin{lstlisting}[caption=Rx Transactions, label=rxtransactions]
Store ID	Product ID	Date	EoD Sales Qty	Qty UoM
\end{lstlisting}

\begin{figure}[!h]
    \centering
	\includegraphics[width=\columnwidth]{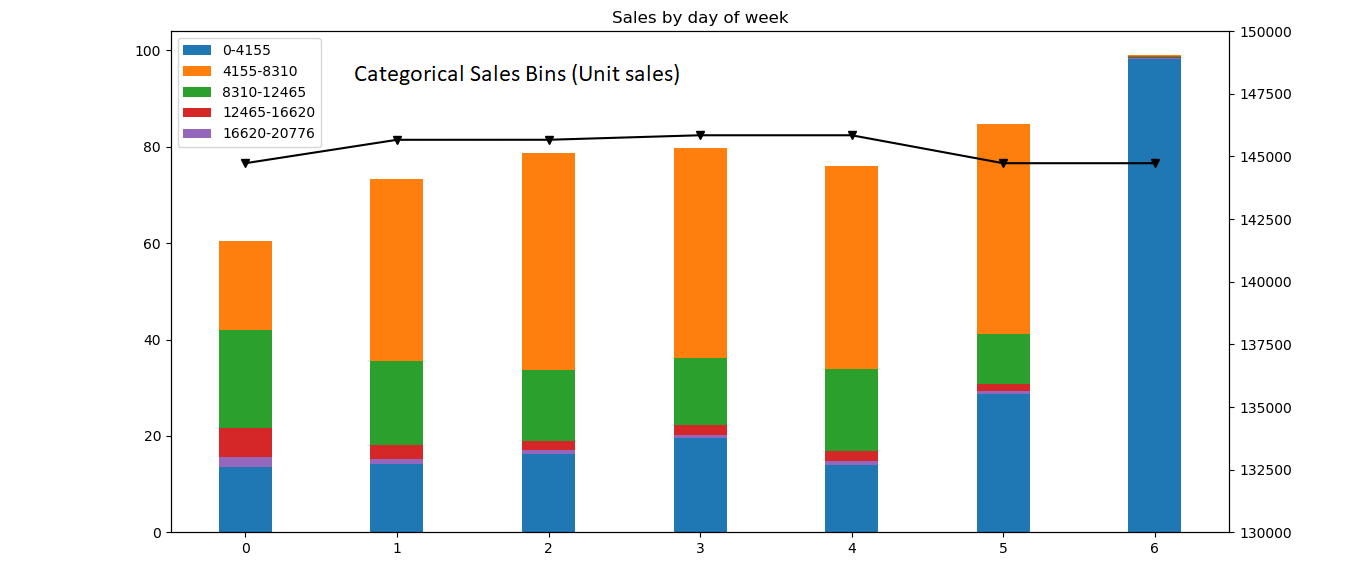}
	\caption{Percentile categorical and total sales by day of the week}
	\label{fig:percentile}
\end{figure}

Figure ~\ref{fig:percentile} shows the sales for five largest categorical sales bins captured by day of the week. Generally there are several thousand data points that need to be analyzed by AI models. Today's AI models are unable to forecast using a lakh or million records and therefore it is a common practice to create subsets of data for modeling. These subsets are referred to as categorical bins; it is like a slice of data representative of the original dataset where the discrete inputs are masked.  And, these categorical bins facilitate the creation of macro models. In this example, we have used 5 categorical bins based on unit sales.

As shown in Figure ~\ref{fig:percentile}, the median sales were higher on the day of a promotional event. It is imperative for retailers to forecast sales accurately to avoid a shortfall in sales on the day of event. This can be complex as store sales vary based on multiple factors including promotions, competition, school and state holidays, seasonality, locality, etc.

\section{Building an RL Model for Forecasting: An Example}

Let's see how a parameterized RL model can be used for accurate demand forecasting.

\begin{enumerate}

\item \textbf{Determine the AI Model: Discrete Vs Continuous:} Considering that the dimensionality of the data sets will grow exponentially, i.e. all permutations and combinations of the goals and the constraints have to be factored, a discrete AI model is best suited. For example, the goals here are the net unit sales for the promotion and the constraints could be any of the parameterized factors such as promotions, seasonality, competition,
and locality.
 
An N-dimensional network is defined using the discrete.DiscreteEnv class. The environment is initialized using a static set of maps that hold the transition probabilities and reward functions. For continuous implementations such as the Cart Pole, use the gym.Env \cite{cartpole}. The code snippet (See Listing ~\ref{frozenlake}) shows how to initialize the Frozen-Lake environment for a run of 100 timesteps. The render function is used in an episode (sequence of events in RL) to visualize the observation space. 

\begin{lstlisting}[caption=Frozen Lake, label=frozenlake]
env = gym.make('FrozenLake-v0')
env.reset()
for _ in range(100):
env.render()
env.step(env.action_space.sample()) # take a random action
env.close()
\end{lstlisting}

\item \textbf{Define the RL Agent:} Next, we define a RL agent that learns the Frozen lake environment based on the observation and reward from episodic events \cite{rlbook}. An episode is a sequence of steps. Actions and rewards are defined for each episode. RL scenarios for an agent in deterministic environment can be formulated as dynamic programming problem. This can be done using the Temporal Difference Learning and Q-Learning functions. RL algorithms consists of online and offline approaches. Online RL algorithms, such as Q-learning explore a domain while learning a policy. Offline approaches use a distinct sample collection phase and learning phases. Each has advantages and disadvantages. Online approaches are computationally efficient and adapt quickly to new observations. More complex real-world problems might utilize an offline algorithm wherein a subset of the data is used to model the actual environment.

In this example, the agent implements a Q-learning based RL step (See Listing ~\ref{qlearning}) to determine the optimal reward function in the customer sales environment. The goal of the agent is to maximize the total reward for any exploratory path in the episode, i.e.  the agent has to perform a series of steps in a systematic manner so that it can learn the ideal solution based on the reward values. 

\begin{lstlisting}[caption=Q-Learning, label=qlearning]
Initialize Q(s,a), $\forall$ s $\epsilon$ S, a $\epsilon$ A(s), arbitrarily, and Q(terminal-state,.) = 0
	Repeat (for each episode):
	Initialize S
	Repeat (for each step of episode):
		Choose A from S using policy derived from Q (e.g., $\epsilon$-greedy)
	Take action A, observe R, S'
	Q(S,A) $\leftarrow$ Q(S,A) + $\alpha$[R + $\gamma$ max Q(S',a) - Q(S,A)]
	S $\leftarrow$ S'
	until S is terminal
\end{lstlisting}

An 'act' function is defined on the agent to determine the appropriate action, followed by a 'step' function to evaluate the outcome of the action in the current environment (see Figure ~\ref{fig:rlagent}). The agent perceives the environment and performs actions and determines the optimal reward in the customer sales environment. 

\begin{figure}[!h]
    \centering
	\includegraphics[width=0.5\columnwidth]{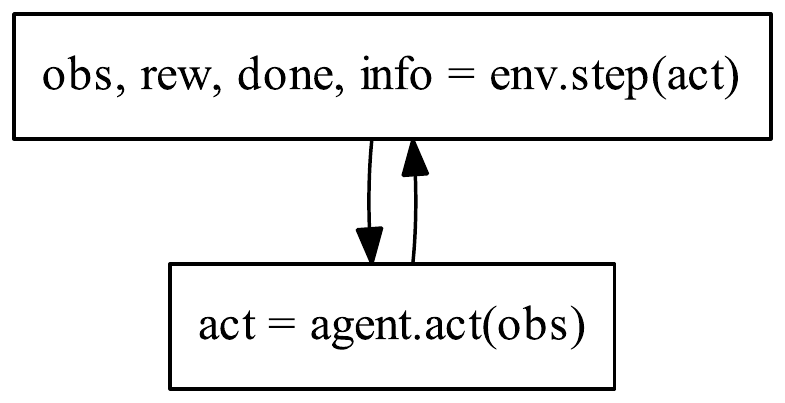}
	\caption{RL agent in an environment}
	\label{fig:rlagent}
\end{figure}

For large state space exploration as in AlphaGo and related DeepMind projects, the agents make use of Deep Learning (DL) techniques \cite{deepgame}.

\item \textbf{Train the RL model:} The model is trained based on key parameters in the dataset such as sales, promo, state holiday, school holiday and day of week. It monitors the sales and adjusts the inventory by using the increase and lower inventory functions. 

If the model is unable to find a promotional channel, it implements a realign operation. This ensures that the episode aligns with the available promotional events as defined in the forecasting dataset; in other words, it implements the business rule to check if there is an active promotion on the channel. This is implemented by using the OpenAI Gym transitions probability matrix discrete.DiscreteEnv.P. The probability matrix P contains the transition mapping for a state and action taken, as well as the possible next states. These are defined using the template in Listing ~\ref{template}.

\begin{lstlisting}[caption=Transition probabilities matrix template, label=template]
{state : { actions : [ (probability, next states, reward, done) ] } }
\end{lstlisting}

\begin{lstlisting}[caption=Transition probabilities, label=probabilities]
  35: { 0: [ (0.14285714285714285, 30, -1, False),
             (0.14285714285714285, 31, -1, False),
             (0.14285714285714285, 32, -1, False),
             (0.14285714285714285, 33, -1, False),
             (0.14285714285714285, 34, -1, False),
             (0.14285714285714285, 35, -1, False),
             (0.14285714285714285, 38, -1, False)],
        1: [(1.0, 25, -1, False)],
        2: [(1.0, 45, -1, False)],
        3: [(1.0, 35, -10, False)]},
  36: { 0: [ (0.14285714285714285, 30, -1, False),
             (0.14285714285714285, 31, -1, False),
             (0.14285714285714285, 32, -1, False),
             (0.14285714285714285, 33, -1, False),
             (0.14285714285714285, 34, -1, False),
             (0.14285714285714285, 35, -1, False),
             (0.14285714285714285, 38, -1, False)],
        1: [(1.0, 26, -1, False)],
        2: [(1.0, 46, -1, False)],
        3: [(1.0, 36, -10, False)]},
\end{lstlisting}

\end{enumerate}

In forecasting promotions, the episode either ends with a promotion or no promotion on a specific day of the week. The next-state feature corresponding to any non-terminal promotional paths (unexpected event in the context of forecasting) are chained to the closest available matches using the input dataset (see Listing ~\ref{probabilities}). 

For example, if the retailer does not run a promotion during a particular week, the model moves to forecast the inventory required for the next event (promotional event or seasonal event).  The model determines the closest available match by using a subset of the matching records to determine the optimal reward at training time for the realign action at each step, i.e. it can either increase, lower, realign, or forecast a promotion. This ensures that the model continuously forecasts inventory irrespective of whether there is an event or not, providing optimal forecasts based on customer data.

\begin{figure}
  \centering
  \includegraphics[width=.8\columnwidth]{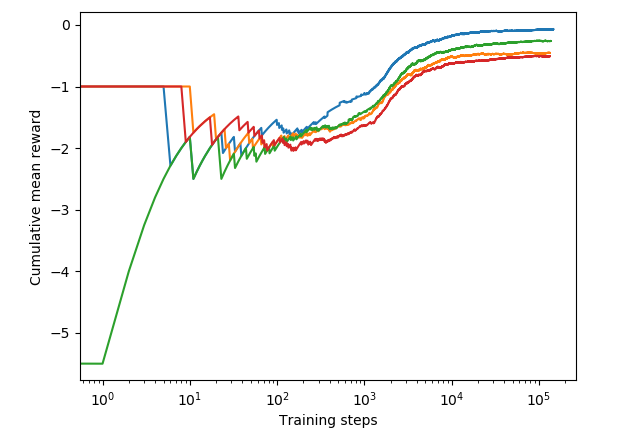}
  \caption{Mean cumulative reward}
  \label{fig:meanreward}
\end{figure}

\begin{figure}
  \centering
  \includegraphics[width=.8\columnwidth]{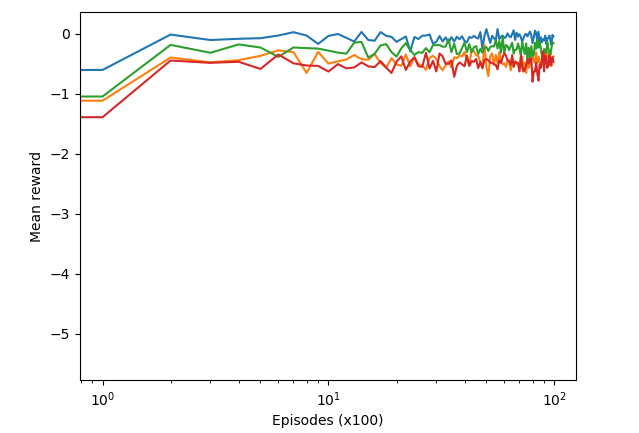}
  \caption{Episodic reward}
  \label{fig:episodicreward}
\end{figure}

Plot in Figure ~\ref{fig:meanreward} shows the mean cumulative reward as a function of the steps for an independent set of episodes. As we see the reward value
gradually saturates in the number of training steps. Figure ~\ref{fig:episodicreward} shows a sample episode for a trained RL model.

\section{Key Findings Indicate Accuracy of RL for Supply Chain Modeling}

\begin{figure}[!h]
\begin{minipage}{.5\textwidth}
    \centering
	\includegraphics[width=.9\textwidth]{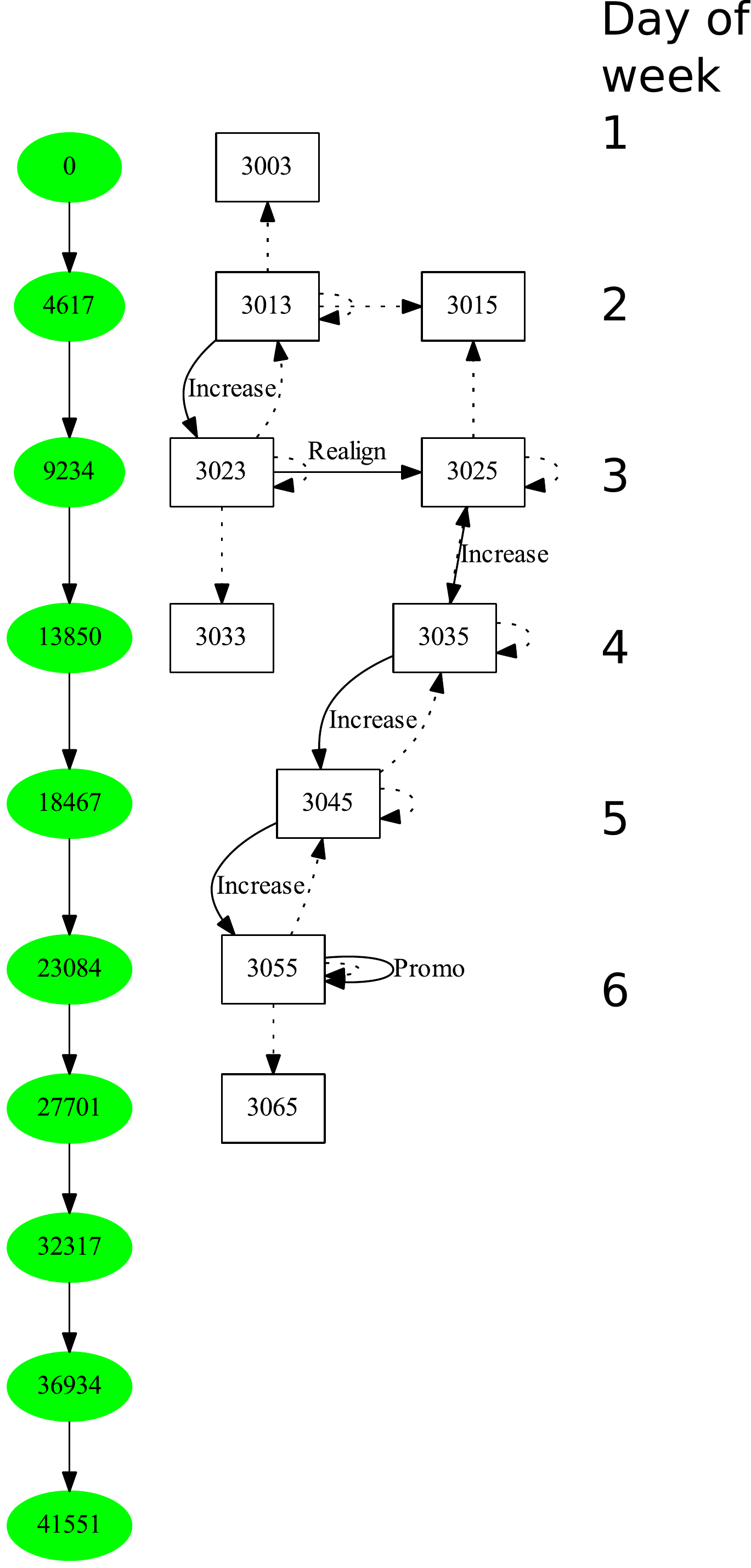}
	\caption{Increase Sales Promotion episode}
	\label{fig:increasesales}
\end{minipage}
\begin{minipage}{0.5\textwidth}
    \centering
	\includegraphics[width=.9\textwidth]{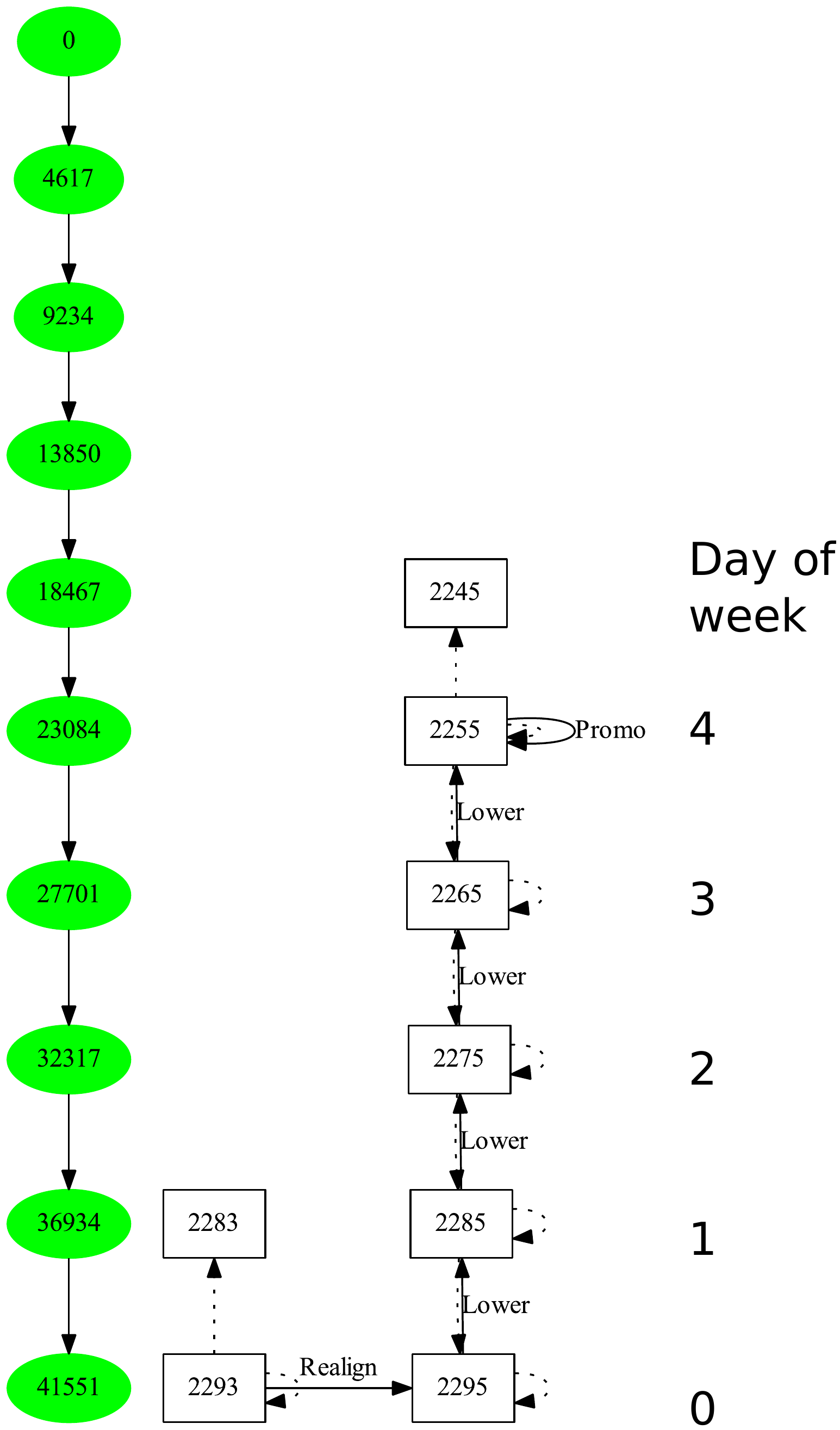}
	\caption{Lower Sales Promotion episode}
	\label{fig:lowersales}
\end{minipage}
\end{figure}

Figures ~\ref{fig:increasesales} and ~\ref{fig:lowersales} show sample episodes. The green circles show the sales captured by the day of the week. We see a steady increase in sales on day-to-day basis leading up to a promotional event. Each set of rows in the figure depict the days of the week corresponding to the episode. Similarly, in Figure ~\ref{fig:lowersales}, we see a lower sales event leading up to a promotion. Realign operations ensure that promotion events for seasonality are forecasted past the attributes in the historical data. The findings indicate that RL models can accurately model forecasting and promotional events corresponding to inventory sales. They capture nuances in the dataset, such as variations in sales from day of week or month of year. 

\section{Conclusion}
We evaluated the suitability of  RL models built with the OpenAI Gym framework for retail SCM. The example of the forecasting model captures forecasting data into a learning algorithm that provides guidance to retailers on the inventory to be stocked in the DC. This model can be tailored to meet the constraints of customer data and advanced use cases in SCM. 

In a discrete RL environment such as the one configured in the example, the model can provide a guide of the possible outcomes in a producer-consumer model using a many-to-one scenario where you could have 4 producers mapped to a single consumer. Because OpenAI  Gym offers a high degree of programmability, enabling both elementary linear maps and higher-dimensional policy functionalities, both simple and sophisticated environments can be modeled to cater to supply chain scenarios.

\section{About the Author}

\textbf{Shaun D'Souza}

Data Scientist, TCS

Shaun has over 12 years of experience in AI, ML, Software Engineering, R\&D, and Business. He has a Bachelor's degree from Cornell University with a Double Major in Computer Science, Electrical and Computer Engineering, Cum Laude Honors, and a Masters in Electrical Engineering from the University of Michigan. Shaun has worked in Software/Business researching machine learning, compilers, algorithms, and systems. He has published several papers and has been granted a patent.

\end{document}